\documentclass[conference]{IEEEtran}
\IEEEoverridecommandlockouts
\usepackage{cite}
\usepackage{amsmath,amssymb,amsfonts}
\usepackage{algorithmic}
\usepackage{graphicx}
\usepackage{textcomp}
\usepackage{xcolor}
\def\BibTeX{{\rm B\kern-.05em{\sc i\kern-.025em b}\kern-.08em
    T\kern-.1667em\lower.7ex\hbox{E}\kern-.125emX}}
    
\usepackage{float}
\usepackage{placeins}
\usepackage{multirow}
\usepackage{graphicx}   
\newcommand*\rot{\rotatebox{90}}

\makeatletter

\def\ps@IEEEtitlepagestyle{%
    \def\@oddfoot{\mycopyrightnotice}%
    \def\@evenfoot{}%
}
\def\mycopyrightnotice{%
    {\footnotesize  978-1-4799-6773-5/14/\$31.00 \textcopyright2021 IEEE\hfill}
    \gdef\mycopyrightnotice{}
}



\makeatletter
\newcommand*\titleheader[1]{\gdef\@titleheader{#1}}
\AtBeginDocument{%
    \let\st@red@title\@title
    \def\@title{%
        \bgroup\normalfont\large\centering\@titleheader\par\egroup
        \vskip1.5em\st@red@title}
}
\makeatother

\titleheader{$36^e$ Congrès de la recherche au CHU Sainte-Justine}

\title{Machine Learning Based on Natural Language Processing to Detect Cardiac Failure in Clinical Narratives\\

}

\author{\IEEEauthorblockN{Thanh-Dung Le}
\IEEEauthorblockA{\textit{Biomedical Information Processing Lab} \\
\textit{École de technologie supérieure}, Canada\\
thanh-dung.le.1@ens.etsmtl.ca}
\and
\IEEEauthorblockN{Rita Noumeir}
\IEEEauthorblockA{\textit{Biomedical Information Processing Lab} \\
\textit{École de technologie supérieure}, Canada\\
rita.noumeir@etsmtl.ca}
\and
\IEEEauthorblockN{Jérôme Rambaud MD PhD}
\IEEEauthorblockA{\textit{CHU Sainte-Justine Hospital} \\
\textit{University of Montreal}, Canada\\
jerome.rambaud-ext@aphp.fr}
\and
\IEEEauthorblockN{Guillaume Sans MD}
\IEEEauthorblockA{\textit{CHU Sainte-Justine Hospital} \\
\textit{University of Montreal}, Canada\\
guillaumesans@orange.fr}
\and
\IEEEauthorblockN{Philippe Jouvet MD PhD}
\IEEEauthorblockA{\textit{CHU Sainte-Justine Hospital} \\
\textit{University of Montreal}, Canada\\
philippe.jouvet@umontreal.ca}

}
\begin{document} 

\maketitle

\begin{abstract}
The purpose of the study presented herein is to develop a machine learning algorithm based on natural language processing that automatically detects whether a patient has a “cardiac failure” or “healthy” condition by using physician notes in Research Data Warehouse at CHU Sainte-Justine Hospital. First, a word representation learning technique was employed by using bag-of-word (BoW), term frequency–inverse document frequency (TF-IDF), and neural word embeddings (word2vec). Each representation technique aims to retain the words’ \textbf{semantic} and \textbf{syntactic} analysis in critical care data. It helps to enrich the mutual information for the word representation and leads to an advantage for further appropriate analysis steps. Second, a machine learning classifier was used to detect the patient’s condition for either cardiac failure or stable patient through the created word representation vector space from the previous step. This machine learning approach is based on a supervised binary classification algorithm, including logistic regression (LR), Gaussian Naive-Bayes (GaussianNB), and multilayer perceptron neural network (MLP-NN). Technically, it mainly optimizes the empirical loss during training the classifiers. As a result, an automatic learning algorithm would be accomplished to draw a high classification performance, including accuracy (acc), precision (pre), recall (rec), and F1-score (f1). The results show that the combination of TF-IDF and MLP-NN always outperformed other combinations with all overall performance. In the case without any feature selection, the proposed framework yielded an overall classification performance with acc, pre, rec, and f1 of 84\% and 82\%, 85\%, and 83\%, respectively.  Significantly, if the feature selection was well applied, the overall performance would finally improve up to 4\% for each evaluation.
\end{abstract}
\IEEEpeerreviewmaketitle

\section{Introduction}
\IEEEPARstart{C}{urrently}, an abundance of data and information are available in the clinical domain. Taking advantage of this opportunity, clinicians have successfully combined the informative and structured data, which includes laboratory test results and medical imaging \cite{abhyankar2014combining}. However, clinical narrative sources are imposed considerable constraints, which are short notes on patients written initially by doctors and physicians. Although the notes are continuously provided and stored in electronic medical records (EMR), they are underutilized in clinical decision support systems. The limitation comes from their unstructured or semi-structured format \cite{johnson2016machine}. 

Since 2013, the Pediatric Critical Care Unit at CHU Sainte-Justine (CHUSJ) has used an EMR. The patients' information, including vital signs, laboratory results, ventilator parameters,is updated every 5 minutes to 1 hour according to the variable sources\cite{matton2016databases}. Mostly, a significant data source of French clinical notes is currently stored. There are 7 caregiver notes/patient/day from 1386 patients (containing a dataset of more than $2.5\times 10^7$ words). These notes are scribed extensively from admission notes and evaluation notes. Admission notes outline reasons for admission to intensive care units, historical progress of the disease, medication, surgery, and the patient's baseline status. Daily ailments and laboratory test results are described in evaluation notes, from which patient condition is evaluated and diagnosed later by doctors. However, these information sources are being used as clinical documentation for reporting and billing instead of prior clinical knowledge for predicting disease conditions. 
\subsection{Problem Statement and Motivation}
One of the targets of the clinical decision support system in CHUSJ is to early diagnosed acute respiratory distress syndromes (ARDS). Theoretically, it endeavors to improve the respiratory disease diagnosis for children by detecting the absence of “cardiac failure” as suggested by a board of experts in pediatric acute lung injury in Table \ref{tab:my_table_ARDS} \cite{pediatric2015pediatric}. Therefore, in such standard criteria for age, timing, the origin of edema, and chest imaging with an acute deterioration in oxygenation can be well explained by underlying cardiac disease. Then, correctly identifying cardiac failure is vital in the diagnosis and treating ARDS. This linked syndrome is diagnosed using many clinical information and exams, including laboratory tests or echocardiography tests. However, the patient cardiac condition can be virtually found based on the stored clinical notes that synthesized all this information. 


Generally, there is a list of golden indicators to classify the patient with cardiac failure. Those indicators could be either from the medical history, clinical exam, chest X-Ray interpretation, recent cardiovascular performance evaluation, or laboratory test results. Medication, such as Levosimendan, Milrinone, Dobutamine, is a surrogate to the gold standard. Its list can be retrieved from syringe pump data, prescriptions, and notes. If any medication from the three is present, there is certainly a cardiac failure. Besides, cardiovascular performance evaluation also highly contributes to indicate the cardiac failure diagnosis. One of the evaluations is ejection fraction (FE) $<50\%$. EF refers to the percentage of blood pumped (or ejected) out of the ventricles with each contraction. It is a surrogate for left ventricular global systolic function, defined as the left ventricular stroke volume divided by the end-diastolic volume. The other indicator for cardiovascular performance evaluation is shortening fraction (FR) $<25\%$. FR is the length of the left ventricle during diastole and systole. It measures diastolic/systolic changes for inter-ventricular septal dimensions and posterior wall dimensions. Finally, brain natriuretic peptide, known as pro-BNP ng/L $>1000$, comes from laboratory test results being useful in the acute settings for differentiation of cardiac failure from pulmonary causes of respiratory distress. Pro-BNP is continually produced in small quantities in the heart and released in more substantial quantities when the heart needs to work harder. 

Consequently, the clinical knowledge representation summarizes detail attributes, which are the main essential contribution to detect cardiac failure, shown in Table \ref{tab:table_pre}. All notes are taken into account if they are encompassed by the information of the prescription history of Milrinone (mcg/kg/min), measurement notes of pro-BNP (ng/L), dilated cardiomyopathy, acute left cardiac failure, chronic cardiac failure, postoperative cardiac failure, coronary microvascular disorder history notes, notes of a measurement result of either FE-ejection fraction(\%) or FR-shortening fraction (\%). As a result, a patient is considered to have a cardiac failure if he/she takes one of the criteria. Unfortunately, as all the mentioned information above that helps diagnose cardiac failure is not readily available electronically, we will develop a machine learning algorithm based on natural language processing (NLP) that automatically detects the desire concept label from clinical notes. Specifically, the algorithm can automatically see whether a patient has a cardiac failure or a healthy condition lacking gold indicators from the notes. Technically, in such a situation, the proposed algorithm can effectively learn a latent representation of clinical notes, which cannot be depicted by traditionally rule-based approaches.

\begin {table*}[!htbp]
\caption{The clinical knowledge representation in detecting cardiac failure \strut}
\label{tab:table_pre} 
\begin{center}
    \begin{tabular}{ | p{15cm} |}
    \hline
     $[\textbf{Label}]$ \\
     \hspace{40pt} Cardiac failure     \\ \hline
    
    $[\textbf{Attributes}]$
     \begin{enumerate}
         \item Admission notes:
            \begin{enumerate}
                \item \textbf{Medical history}: Levosimendan, Milrionone or Dobutamine.
                \item \textbf{History of diagnosed terms}: cardiomyopathie dilatee, choc cardiogenique, defaillance cardiaque gauche aigue, defaillance cardiaque gauche chronique, defaillance cardiaque post-operatorie (LCOS), surcharge liquidienne (hypervolemie), myocardite.
            \end{enumerate}
         \item Evaluation notes:
            \begin{enumerate}
                \item \textbf{Evolution par systeme} (Cardiovascular): FE (ejection fraction $<$ 50\%) and/or FR (shortening ratio $<$ 25\%).
                \item \textbf{Laboratory test result}: pro-BNP ng/L ($>$ 1000)
            \end{enumerate}
     \end{enumerate}
     \\ \hline
    
    \end{tabular}
    
\end{center}
\end{table*}

\subsection{Related Works}

Within a decade, there is a 9 times increase in EHR system adoption in hospitals in 2018 compared to 2008 \cite{jiang2017artificial}. Because of increasing EMR, the stored data in EMR also proportionally increased, however, not all data is directly used to analyze health management improvement. The data must be preprocessed before use, and there are different pre-processing methods in conjunction with varying limitations for various data types. For example, clinical notes have a lack of standardization \cite{kreimeyer2017natural}. Therefore, it increases the complexity of the knowledge extraction from clinical text data. That is why there is limited clinical text adoption in real practice, although clinical notes are holding a developing position in clinical research priority.

Currently, there has been a outstanding achievement of machine learning-based NLP for clinical text knowledge extraction. However, remained significant concerns are being the trajectory of clinical NLP research discussed in \cite{young2018recent} and \cite{sheikhalishahi2019natural}. Especially in cases of the clinical notes in language other than English, the challenge is more difficult to solve \cite{neveol2018clinical}. First, one remained limitation is how to learn the semantic and syntactic structure of feature extraction for clinical texts. The information combination (semantic and syntactic structure) will lead to conflation deficiency when a word level is tightened to the semantic and syntactic level. Hence, the target is cognizance enrichment during the combination process instead of losing crucial information. Second, supervised learning algorithms for classification have reached their highest potentiation because they only learn the discrepancy between instances from actual data distribution and synthesized instances generated by the learning rules. However, the underlying probability distribution of clinical text is not known; thus, the solution to the classification is deficient.

Therefore, as proven in the recent study \cite{olsen2020clinical}, the feasibility of employing machine learning for cardiac failure was extensively analyzed and confirmed. However, in our case, we are dealing with two challenge: (i) clinical notes in French, and (ii) a limited amount of dataset size. And, it will also be attainable by effectuating specific questions we aim to answer:
\begin{itemize}
    \item  Which representation learning approach should be used? The representation learning approach, which can retain  the  words’  semantic  and  syntactic  analysis  in critical care data, enriches the mutual information for the word representation by capturing word to word correlation not only from statistically but also semantically distributional learning of word appearance.
    \item  Which machine learning classifier should be employed? The machine classifier, which can  influentially  avoid  the  overfitting associated  with  machine  learning  rule  by  marginalizing  over the  model  parameters  instead  of  making  point  estimates  of its  values, would be assuredly accomplished to draw a high classification accuracy, recall, and precision for the desire labeling concept.
\end{itemize}

The next section \ref{sec:data_prep} will discuss the data and methodology. First, we will discuss data preparation, including data collection, labelling and pre-processing from CHUSJ. Then, we will explore the possibilities of various clinical note representation learning techniques as Bow, TF-IDF, and neural word embeddings. Following, different machine learning classifiers performance is compared with regard to LR, GaussianNB and MLP-NN. The experimental results will then be discussed in section \ref{sec:results}. Finally, section \ref{sec:conclusion} provides concluding remarks.

\section{Materials and Methods}
\label{sec:data_prep}
\begin{figure}[t]
			\centering
			\includegraphics[width=0.48\textwidth]{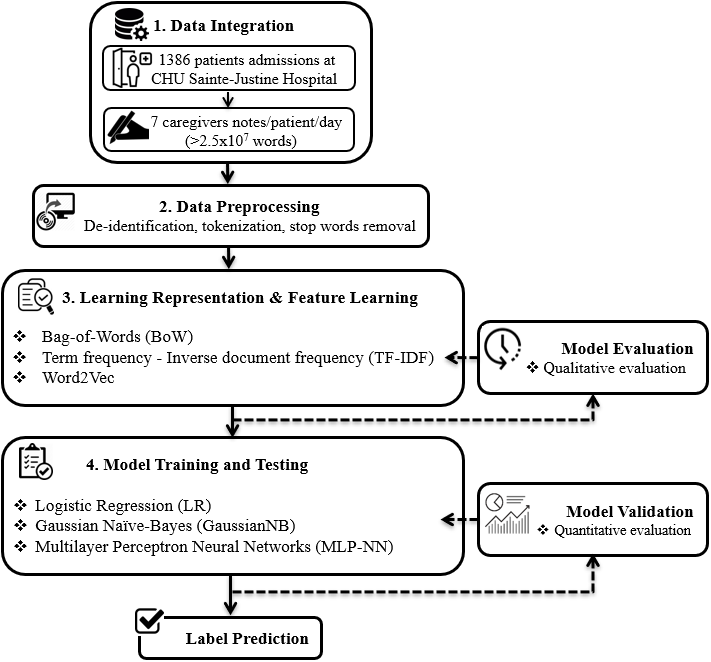}
			\caption{An overview of the proposed methodology to detect cardiac failure from clinical notes at CHUSJ.}
			\label{fig:method}
		\end{figure}

The principal methodology is empirical experiments of a learning algorithm to learn the hidden interpretation and presentation of the French clinical note data. Technically, an optimal learning algorithm will be achieved by minimizing the empirical loss during the learning process of supervised learning. Generally, a detection algorithm based on supervised learning follows the same four steps, including data integration, data pre-processing, feature extraction, classification, and prediction. Nevertheless, the performance of the detection techniques strongly depends on the details of each step above. As shown in Fig. \ref{fig:method}, it is divided into four steps as follows:
\begin{enumerate}
    \item First, categories of clinical notes, including admission and evaluation notes from CHUSJ databases are collected in the data integration process. 
    
    \item Next, data pre-processing will be performed to pre-process raw text data. Notably, the information of the data is tightened by applying lowercase conversion, stop words removal and  tokenization. 
    
    \item Followed the pre-processing step, a representation learning model is built-in feature extraction. The model is trained to compress a text sequence into a low-dimensional vector space, learn the word representation, and integrate relation from lexical resources into the learning process. 
    
    \item Finally, machine learning classifier is iteratively trained to learn the knowledge representation of processed data in preceding steps. Consequently, the final result must obtain high accuracy in cardiac failure prediction. 

\end{enumerate}

\subsection{Clinical Narratives Datasets}

The data integration process has been successfully completed at the Pediatric Intensive Care Unit, CHUSJ, for more than 1300 patients. More than 5000 single line of notes. We only took information from two types of notes, including admission and evaluation notes. Primarily, we focused on medical background, history of the disease to admission, and cardiovascular evaluation. Furthermore, we only used notes for each patient's first stay within the first 24h since the admission. If a patient had more than one ICU stay, we only analyzed the first one. Two doctors from the CHUSJ (Dr. Jérôme Rambaud and Dr. Guillaume Sans) separately reviewed each patient's notes; then, each note was manually labeled "Positive" or" Negative" for cardiac failure or healthy condition, respectively. Finally, to avoid data contamination, we checked both the patientID and careproviderID to ensure that there were not any notes being present in the training cohort and the testing cohort at the same time.

\begin{figure*}[t]
	\centering
	\includegraphics[scale=0.64]{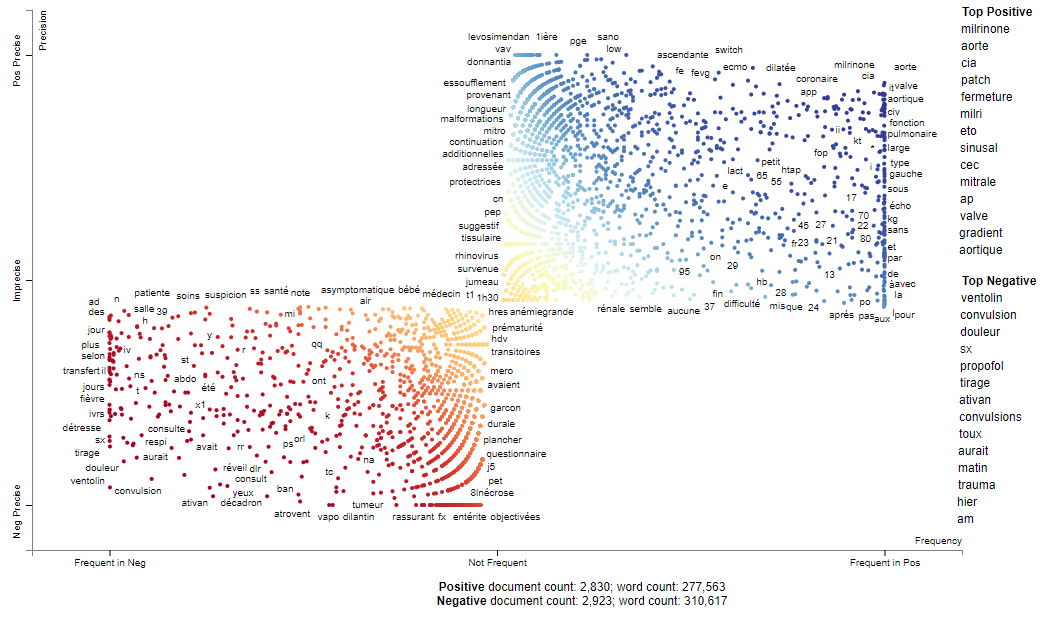}
	\caption{Clinical note illustration for both positive and negative cases of cardiac failure at CHUSJ.}
	\label{fig:clinical_nlp_illustration}
\end{figure*}

\subsection{Data Preprocessing}
Generally, it is proven that if the preprocessing steps are well equip, the result for the end-task will be improved \cite{kannan2014preprocessing}. Therefore, there are steps that were used as case lowering, and stop words removing. Finally, we tokenized the narratives into a vector of tokens. Totally, we had more than 580000 word count from the data shown in Fig. \ref{fig:clinical_nlp_illustration}. The list of top words from the positive and negative cases was also illustrated on the right-hand side. Since then, we quickly see that all the terms following are positively related to the cardiac malfunction: milrinone (milri), aortique, (aorte), cia, cec, valve, etc. These terms are summarized in Table \ref{tab:medical_abbreviation}, and they are compatible with the clinical diagnosis mentioned from Table \ref{tab:my_table_ARDS}, \ref{tab:table_pre}.

\begin{table*}[]
\centering
\caption{Important Abbreviations for Medical Terms}
\label{tab:medical_abbreviation}
\begin{tabular}{|l|l|l|}
\hline
\multicolumn{1}{|c|}{Abbreviations} & \multicolumn{1}{c|}{Descriptions}             & \multicolumn{1}{c|}{Characteristics}       \\ \hline
CIV & Communication intraventriculaire & Cardiac malformation          \\ \hline
CEC & Circulation extracorporelle      & Treatment for cardiac failure \\ \hline
CIA & Communication intraauriculaire   & Cardiac malformation          \\ \hline
FC  & Fréquence cardiaque              & Cardiac frequency             \\ \hline
IVRS                                & Infection des voies respiratoires supérieures & Virus responsible for respiratory distress \\ \hline
SOP & Salle d'opération                & Operations                    \\ \hline
PO  & Per os (by mouth)                & Feeding                       \\ \hline
\end{tabular}
\end{table*}

\subsection{Feature Learning}
\subsubsection{Bag-of-Words}

A bag-of-words model, or BoW for short, is a representation of text that describes the occurrence of words within a document. Technically, documents have similar content, and they will be similar. It extracts features from the text by considering each word count as a feature from a vocabulary of known words. There is a theoretical analysis for understanding the BoW model is presented \cite{zhang2010understanding}. 

\subsubsection{TF-IDF}
The TF-IDF, first introduced in \cite{salton1973specification}, stands for term frequency (TF) × inverse document frequency (IDF). TF-IDF weighting is commonly used in information retrieval for text mining. The intuition is that term importance increases with the term's frequency in the text, but the term's frequency neutralizes it in the domain of interest. A detailed explanation of the TF-IDF model based on a probabilistic justification is discussed in \cite{ havrlant2017simple} for its heuristic and other variations.






\subsubsection{Neural Word Embeddings}




The neural word embedding was introduced in \cite{mikolov2013distributed}, and named word2vec model. The central target word vector and context word vector for each word are the neural word embedding parameters. The key to the success of word2vec is that it can compute the logarithmic conditional probability for the central word vector and the context word vector. Consequently, its computation obtains the conditional likelihood for all the words in the dictionary given the central target word  $w_c$, and trained by a neural network. 

\begin{align}
    \log P(w_o \mid w_c) =
\mathbf{u}_o^\top \mathbf{v}_c - \log\left(\sum_{i \in \mathcal{V}} \text{exp}(\mathbf{u}_i^\top \mathbf{v}_c)\right)
\end{align}




\subsection{Machine Learning Classifiers}
Although deep learning has proven its superiority to representation learning and classification problems from various data structures such as medical imaging, time series data, and clinical natural language. Unfortunately, not all cases can apply deep learning, especially with limited data \cite{paleyes2020challenges}. Therefore, in this study, we decide to go with three different machine learning classifiers logistic regression (LR), Gaussian Naive-Bayes (GaussianNB), and multilayer perceptron neural network (MLP-NN) because of their simplicity.

\subsubsection{Logistic Regression (LR)}

LR uses a logistic function to model the probability of a  binary dependent variable (particular class or event existing) such as unhealthy/healthy. Therefore, it is widely used in most medical fields \cite{tolles2016logistic}.



\begin{align}
    p(x ; w)=\operatorname{Pr}(y=1 \mid x ; w)=\frac{1}{1+\exp \left(-w^{T} x\right)}
\end{align}

where $w$ is the weight vector of coefficients, and $p()$ is a sigmoid function. Here we assume that the $n$ training examples are generated independently. 


\subsubsection{Gaussian Naive Bayes (GaussianNB)}
GaussianNB algorithm for classification is a set of supervised learning algorithms based on applying Bayes’ theorem with the “naive” assumption of conditional independence between every pair of features given the value of the class variable. However, the difference between GaussianNB and naive Bayes classifiers is mainly based on distribution of $P(x_i \mid y)$. The likelihood of the features is assumed to be Gaussian for GaussianNB \cite{hastie2009elements}: 










\begin{align}
    P(x_i \mid y) = \frac{1}{\sqrt{2\pi\sigma^2_y}} \exp\left(-\frac{(x_i - \mu_y)^2}{2\sigma^2_y}\right)
\end{align}

\subsubsection{Multilayer Perceptron Neural Network (MLP-NN)}
A commonly used neural network is the MLP-NN. In an MLP-NN, the neurons are structured into layers, consisting of at least three layers, namely the input layer, hidden layer or layers, and an output layer \cite{demuth2014neural}. Typical MLP-NN networks are feedforward neural networks where the computation is carried out in a single direction from input to output. It is proven that rectified linear unit (ReLU) activation in combination with stochastic gradient descent shows a better convergence and gives a better optimization in small MLP-NN \cite{li2017convergence}. Therefore, we apply the ReLU activation as shown below:

\begin{align}
    ReLU(x) = \max(x, 0)
\end{align}





\section{Experimental Results and Discussions}
\label{sec:results}
		
To implement, we used the scikit-learn
library (version 0.24.1 \cite{scikit-learn}) in Python. To make our results more consistent, we used the $k$-fold cross validation; each dataset was divided into $k$ subsets called folds, the model was trained on $k-1$ of them and tested on the last left out. This process was repeated $k$ times and the results were averaged to get the final results. Furthermore, we also employed the univariate feature selection with sparse data from the learning representation feature space. This selection process works by selecting the best features based on univariate statistical tests named SelectKBest algorithms, which removes all but the  $K$ highest scoring features.
		
Then, the presented approach could directly be applied to predict the cardiac failure. As shown in Table \ref{tab:table-results}, the results showed that the combination of TF-IDF and MLP-NN always outperforms other combinations with all overall performance. In the case without any feature selection, the proposed framework yielded an overall classification performance with acc, pre, rec, and f1 of 84\% and 82\%, 85\%, and 83\%, respectively.  

Besides, with the same learning presentation approach (BoW, TF-IDF, or embeddings), the LR classifiers had better performance than GaussianNB classifiers. However, MLP-NN models always dominated with their best generalization.

Furthermore, with the limited amount of data, the BoW and TF-IDF were proven their capacity to retain better for the information retrieval in clinical note representation. While the neural word embeddings also showed their potentiality dealing with learning representation for clinical notes. Although the difference between these performances was not significant, it will be more effective when the amount of data is scale-up.

Also, with the high sparsity ratio from the representation feature space, if the feature selection (SelectKBest) was well applied and tuned, it could accomplish the improvement up to 4\% for each evaluation in the overall performance acc, pre, rec, and f1 at 87\% and 86\%, 88\%, and 87\%, respectively.

Finally, it is necessary to enable safe clinical deployment when the uncertainty can be effectively managed. From the preliminary results above, there are two limitations for the model, which should be improved in future works. First, instead of removing all of the vital sign numeric value, we can take advantage of informative relation from those numeric values for the learning representation feature space. And it could be done by deploying different decoding approaches. Second, it has been proved that the sparsity reduction for the feature space strongly affects the task-end classifier performance. Therefore, an advanced learning algorithm is then proposed so that it can effectively leverage the sparsity reduction.


\begin{table}[t!]
\centering
\caption{Summarization of experiments performance evaluation}
\label{tab:table-results}
\begin{tabular}{|c|c|l|c|c|c|c|}
\hline
\multicolumn{2}{|c|}{Representation}                     & ML Approach       & ACC  & PRE  & REC  & F1   \\ \hline
\multirow{9}{*}{\rot{W/o Feature Selection}} & \multirow{3}{*}{BoW} & LR & 0.81 & 0.80 & 0.82 & 0.81 \\ \cline{3-7} 
 &                             & GaussianNB & 0.78 & 0.76 & 0.81 & 0.78 \\ \cline{3-7} 
 &                             & MLP-NN   & 0.81 & 0.80 & 0.82 & 0.81 \\ \cline{2-7} 
 & \multirow{3}{*}{TF-IDF}     & LR  & 0.78 & 0.76 & 0.79 & 0.77 \\ \cline{3-7} 
 &                             & GaussianNB & 0.77 & 0.74 & 0.80 & 0.77 \\ \cline{3-7} 
 &                             & MLP-NN   & \textbf{0.84} & \textbf{0.82} & \textbf{0.85} & \textbf{0.83} \\ \cline{2-7} 
 & \multirow{3}{*}{Embeddings} & LR  & 0.76 & 0.74 & 0.79 & 0.76 \\ \cline{3-7} 
 &                             & GaussianNB & 0.76 & 0.71 & 0.79 & 0.75 \\ \cline{3-7} 
 &                             & MLP-NN   & 0.77 & 0.76 & 0.78 & 0.77 \\ \hline
\multirow{9}{*}{\rot{W/ Feature Selection}}    & \multirow{3}{*}{BoW} & LR & 0.81 & 0.81 & 0.79 & 0.80 \\ \cline{3-7} 
 &                             & GaussianNB  & 0.80 & 0.78 & 0.79 & 0.78 \\ \cline{3-7} 
 &                             & MLP-NN   & 0.82 & 0.82 & 0.81 & 0.81 \\ \cline{2-7} 
 & \multirow{3}{*}{TF-IDF}     & LR  & 0.82 & 0.81 & 0.83 & 0.82 \\ \cline{3-7} 
 &                             & GaussianNB & 0.81 & 0.82 & 0.79 & 0.80 \\ \cline{3-7} 
 &                             & MLP-NN   & \textbf{0.87} & \textbf{0.86} & \textbf{0.88} & \textbf{0.87} \\ \cline{2-7} 
 & \multirow{3}{*}{Embeddings} & LR  & 0.80 & 0.79 & 0.80 & 0.79 \\ \cline{3-7} 
 &                             & GaussianNB & 0.79 & 0.79 & 0.79 & 0.79 \\ \cline{3-7} 
 &                             & MLP-NN   & 0.80 & 0.80 & 0.80 & 0.80 \\ \hline
\end{tabular}
\end{table}

\section{Conclusion}
\label{sec:conclusion}
This paper has employed both learning representation and machine learning algorithms to tackle the French clinical natural language processing for detecting a cardiac failure in children at CHUSJ. Our numerical studies have confirmed that the feasibility of the proposed design by combining TF-IDF and MLP-NN compared to others scenario can be achieved; the proposed mechanism could also be improved with the feature selection from the learning representation vector space.

In the current work, we assumed that the vital sign numeric values are not significant contribution to the classifier. Instead of losing them, we plan to address the different decoding approaches for numeric values in our future work. Furthermore, advanced compressed sensing techniques for sparsity reduction in learning representation space will be carefully considered.

\section*{Acknowledgment}
This work was supported in part by the Natural Sciences and Engineering Research Council (NSERC), in part by the Institut de Valorisation des données de l’Université de Montréal (IVADO), in part by the Fonds de la recherche en sante du Quebec (FRQS), and in part by the Fonds de recherche du Québec – Nature et technologies (FRQNT).

\bibliographystyle{IEEEtran}
\bibliography{IEEEabrv,Bibliography}
\vspace{12pt}

\end{document}